\documentclass{article}
\usepackage{spconf,amsmath,graphicx}

\usepackage{hyperref}
\usepackage{overpic}
\usepackage{multirow}
\usepackage{placeins}
\usepackage[caption=false]{subfig}

\title{Mapping EEG Signals to Visual Stimuli: A Deep Learning Approach to Match vs. Mismatch Classification}
\name{Yiqian Yang{$^1$}, Zhengqiao Zhao{$^1$}, Qian Wang{$^2$}, Yan Yang{$^1$}, Jingdong Chen{$^1$}, \textit{FELLOW, IEEE}}
\address{{$^1$} Center of Intelligent Acoustics and Immersive Communications\\
Shaanxi Provincial Key Laboratory of Artificial Intelligence\\
School of Marine Science and Technology\\
Northwestern Polytechnical University, Xi’an, Shaanxi 710072, China \\
{$^2$} North Electro-Optic Science \& Technology Defense Co.Ltd}

\begin{document}
\maketitle
\begin{abstract}
{Existing approaches to modeling associations between visual stimuli and brain responses are facing difficulties in handling between-subject variance and model generalization. Inspired by the recent progress in modeling speech-brain response, we propose in this work a ``match-vs-mismatch'' deep learning model to classify whether a video clip induces excitatory responses in recorded EEG signals and learn associations between the visual content and corresponding neural recordings. Using an exclusive experimental dataset, we demonstrate that the proposed model is able to achieve the highest accuracy on unseen subjects as compared to other baseline models. Furthermore, we analyze the inter-subject noise using a subject-level silhouette score in the embedding space and show that the developed model is able to mitigate inter-subject noise and significantly reduce the silhouette score. Moreover, we examine the Grad-CAM activation score and show that the brain regions associated with language processing contribute most to the model predictions, followed by regions associated with visual processing. These results have the potential to facilitate the development of neural recording-based video reconstruction and its related applications.}
\end{abstract}
\section{Introduction}
\label{sec:intro}
The ability to predict the brain's neural representation in response to external stimuli holds great potential for advancing fundamental neuroscience research and helping develop key components of Brain-Computer Interfaces (BCI). Thanks to recent advancements in large-scale neural recording such as electroencephalography (EEG), wide-field calcium imaging, and magnetic resonance imaging, scientists can now computationally learn the complex relationship between animal behavior and the corresponding neural activities \cite{urai2022large}. To study neural dynamics in a controlled manner, researchers present visual stimuli such as images and videos to subjects and synchronously record their neural signals. Machine learning models are subsequently trained to learn the relationship between neural signals and stimuli, enabling the reconstruction of visual content from neural recordings. 

Prior studies focused on analyzing spectral signatures of neural recordings to comprehend neural firing mechanisms \cite{ray2011different,rich2017spatiotemporal}, and explored the use of traditional frequency features to reconstruct both overt and imagined speech \cite{DecodingImaginedSpeech_EEGproix2022imagined}. Recent progress in deep learning has allowed researchers to delve into more complex feature spaces of neural signals and nonlinear relationships between external stimuli and their corresponding neural representations. It is demonstrated that deep generative models can be trained to recover category-level images from EEG signals \cite{EEG_image_reconstruction_bai2023dreamdiffusion}. Furthermore, a recent work demonstrated the possibility of reconstructing video frame order from the visual cortex recordings using a contrastive learning-based model  \cite{cebra_nature_schneider2023learnable}. MinD-Video was proposed to extract semantic features from time-continuous fMRI signals of the cerebral cortex and restore semantically accurate videos \cite{mind_video_fmri_chen2023cinematic}.
Nevertheless, the aforementioned reconstruction approaches still face great challenges due to two primary factors. First, the neural response latency is correlated with the subject's attention level due to ``attention drift'', and a large within-subject variance is often observed in repeated experiments \cite{wong2018accurate}. How to extract invariant neural signal patterns that correspond to external stimuli remains a significant challenge. Second, there are large differences in neural response signals between different individuals \cite{mind_video_fmri_chen2023cinematic}. Therefore, models obtained on training subjects may fail to generalize to holdout subjects \cite{cross-subject2023}. To address these challenges, researchers have reframed the EEG-stimuli mapping problem as one of "match-vs-mismatch" classification instead of one of regression, leading to the successful reconstruction of the auditory envelope from EEG signals \cite{wong2018accurate,match_mismatch_speech_eeg_dcnn_accou2021modeling}.

Inspired by the ideas in  \cite{wong2018accurate,match_mismatch_speech_eeg_dcnn_accou2021modeling}, we propose in this work an approach to learning the neural representation of visual content through a ``match-vs-mismatch'' framework. Briefly, we develop a deep learning-based classification model to predict whether a video segment evoked the query segment of the EEG signal. Existing ``match-vs-mismatch'' models typically leverage convolutional neural network (CNN) \cite{CNN_lecun1989backpropagation} and dilated convolutional neural network (DCNN) \cite{DCNN_dilated_convolution_yu2015multi} to extract the local features as well as the global patterns of neural signals. In contrast, we explore in this work the use of sequence models, such as Transformer (TRFM) \cite{transformer_vaswani2017attention}, and Gated Recurrent Unit (GRU) \cite{GRU_chung2014empirical}, in addition to convolutional layers, to capture the contextual information of videos and neural recordings. We will demonstrate the benefits of using sequence models over the convolutional layers-based baseline models. We further analyze the developed models by extracting intermediate layer output and visualizing the channel importance scores. It is shown that the proposed models can capture meaningful neural representations in the embedding space and mitigate inter-subject variance. The major contributions of this work are as follows. 1) We propose to model the visual stimuli and neural response using a ``match-vs-mismatch'' framework. 2) A CNN and GRU based-classification model is developed to predict the video that evoked the input EEG signal, which is able to produce a significantly better performance as compared to other competing models; 3) Various visualization methods are applied to analyze the developed model and the results show that the proposed model is capable of capturing neurologically meaningful features while effectively handling inter-subject noise.

\section{Methods}
\label{sec:methods}

\begin{figure}[!htp]
    \centering
    \includegraphics[width=1\linewidth]{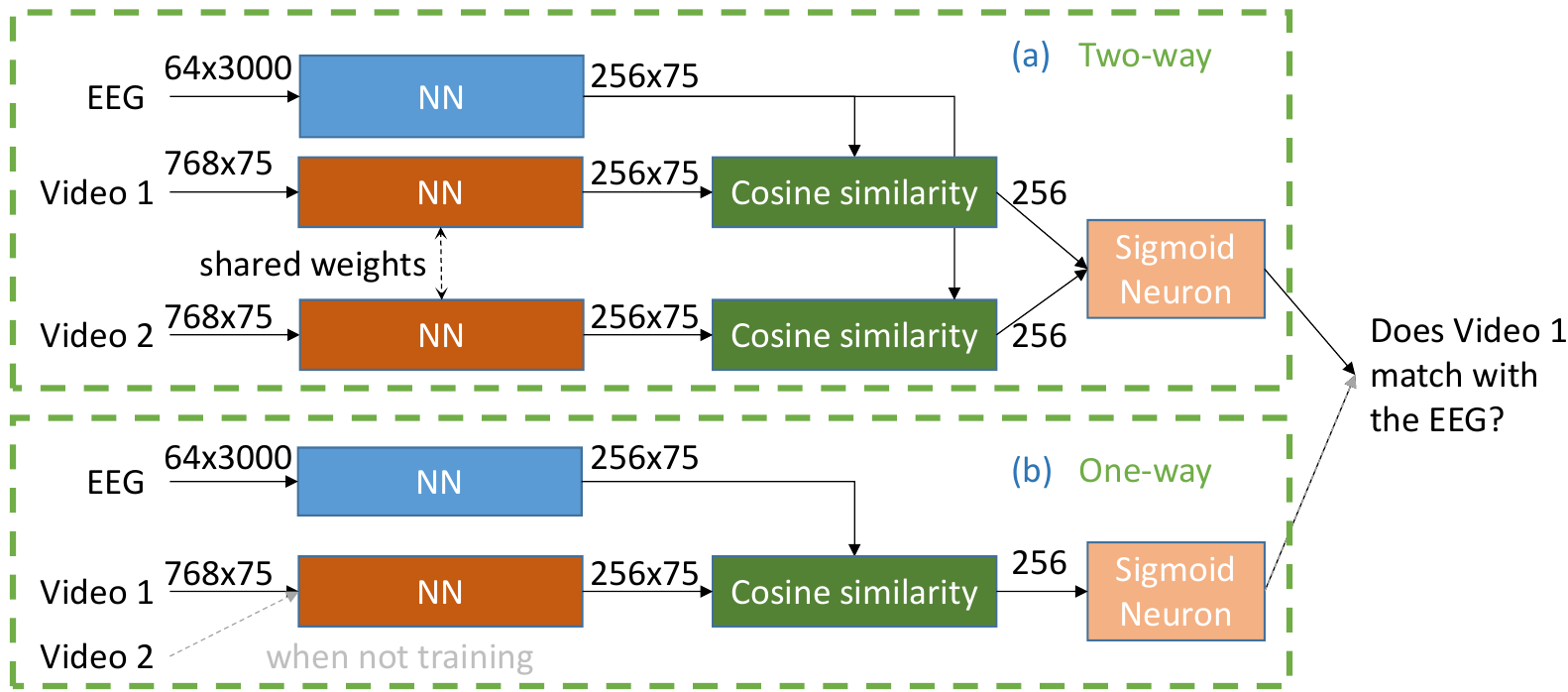}
    \caption{Illustration of the ``match-vs-mismatch'' models: (a) the two-way model and (b) the one-way model.}
    \label{fig:overall_framework}
\end{figure}

The classification model in this work consists of two branches of inputs: 1) the EEG branch and 2) the video branch, as shown in Fig.~\ref{fig:overall_framework} (a). The former takes a segment of the EEG signal as its input while the latter takes two video segments as its inputs. Specifically, the corresponding ``matching'' stimulus video that evokes the EEG signal and a ``mismatching'' imposter video are randomly assigned to video branch input ports 1 and 2. The classification task is to predict whether video 1 is the ``matching'' video. Note that this framework is inspired by previous studies on modeling brain responses to audio stimuli \cite{wong2018accurate,match_mismatch_speech_eeg_dcnn_accou2021modeling}, which we refer to as a two-way classification problem. To verify the benefit of this two-way classification problem formulation, we develop a one-way classification model as a baseline model, where the video branch takes only one video segment as input and determines whether this input matches with the EEG signal, as shown in Fig.~\ref{fig:overall_framework} (b). In both models, the EEG and video signals are processed by different deep neural networks respectively (i.e., the NN blocks in Fig.~\ref{fig:overall_framework}). The cosine similarity is computed between the EEG feature and the video feature per channel along the time dimension. The similarities for different channels are then concatenated. Finally, a fully connected layer with sigmoid activation is used to predict whether the video input matches with the EEG signal. For the two-way model, the networks used to process video input 1 and input 2 share the same weights.

Although most existing works use convolutional layers in the NN blocks to extract features from the EEG and video signals, we believe that sequence models can help learn the contextual information, thereby improving the performance. Therefore, we develop several models that consist of a combination of deep neural network modules such as CNN, DCNN, GRU, TRFM, and LSTM \cite{LSTM_hochreiter1997long}, and compare them to baseline models. Note that in this study we also use an architecture similar to the convolutional networks proposed in \cite{match_mismatch_speech_eeg_dcnn_accou2021modeling} as another baseline model in addition to the aforementioned one-way baseline model. The network architecture of the one-way baseline model is the same as the best-performing architecture of the two-way models.

For models with DCNN, the upstream CNN layers simply perform point-wise convolutional operations, which can keep the same convolution channels as the input channels for downstream layers. For models without DCNN, both the kernel size and the stride size are 40 for the CNN layers. In DCNN, there are three dilated layers with an output dimension of 256 and a kernel size of 5. The dilation factor is chosen to be $k^{n}$, where $n$ is the layer index of dilated layers starting from $0$, and padding size is calculated as $floor(((k-1)\times k^{n}+1)/2)$ in order to keep the temporal dimension divisible by the stride.  A rectified linear unit (ReLU) non-linearity is applied after each dilated convolution layer. We choose the hyperparameters of the networks such that the EEG signals and the video signals are compatible in the temporal dimension. For sequence models, We adopt TRFM modules to capture the potential long-term dependencies. The TRFM module consists of 3 layers with a single attention head of 256 or 768 dimensions and a dropout rate of $0.2$. GRU and LSTM modules consist of 256 hidden neurons. 

\section{Experiments}
\label{sec:evaluation}
We have developed a comprehensive database, which consists of 64-channel EEG recordings obtained from a cohort of 100 Chinese participants while watching a 3.5-minute Chinese movie clip. The sampling frequency for the EEG signal is 1000 Hz and the video has a resolution of 1080p with a frame rate of 25 frames per second. In this study, a subset of the dataset is used to evaluate the proposed models (currently available for research use upon request). In this experimental subset, there are 56 subjects. The total length of the EEG signal from each subject is 210000 samples, and the length of the corresponding video signal is 5250 frames. We use the MNE toolbox to preprocess the EEG signals \cite{mne_toolbox_esch2019mne}. Briefly, we apply a notch filter to mitigate the impact of the power-line interference at 50 Hz. Then, a band-pass filter is applied, which has a passband ranging from $1$ to $200 \mathrm{Hz}$. Moreover, we normalize the maximum magnitude of the EEG signal to 0.8 to ensure the model stability. For video input, we use a pretrained ViT-B/8 model, DINO \cite{dino_v1_caron2021emerging}, to preprocess the video and extract the preliminary feature matrix from the video clip. 

\begin{figure}[!htp]
\centering
	\subfloat[train-test split. \label{fig:datasets_configuration}]{\includegraphics[width = 0.45\linewidth]{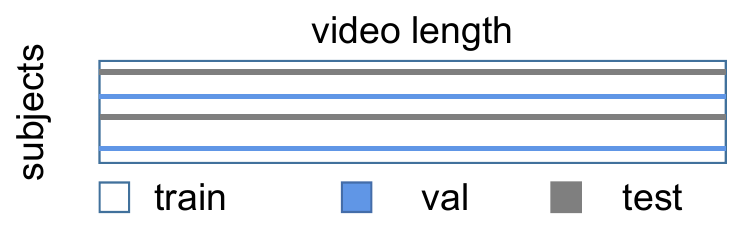}}\hskip 5pt
	\subfloat[matching \& mismatching sampling strategy.\label{fig:match_mismatch_eeg_and_video}]{\includegraphics[width = 0.45\linewidth]{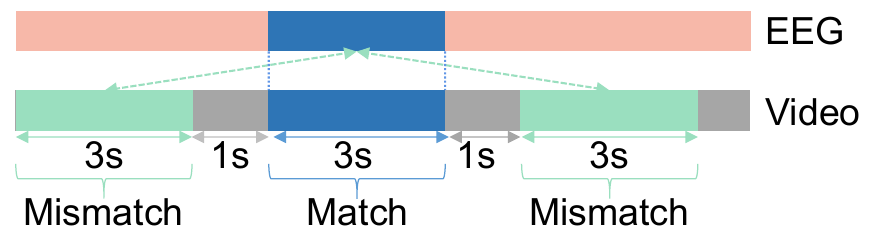}}
	\hfill
\caption{Illustration of the experimental data setup.}
\label{fig:acc_with_separation_secs}
\end{figure}

Data are fed to the models on a segment basis with a segment size of 3 seconds and a shift of 1 second between consecutive segments. The start time of the imposter video segment is chosen exactly 1 second after the end (positively shifted samples) or 4 seconds before the start of the current EEG segment (negatively shifted samples), as illustrated in Fig.~\ref{fig:match_mismatch_eeg_and_video}. The corresponding video segment is preprocessed by the DINO model to produce a 768-dimensional vector per frame. The video feature matrix input for a 3-second video clip is 768 by 75 as shown in Fig.~\ref{fig:overall_framework}. We randomly choose 45, 5, and 6 subjects (18630, 2070, and 2480 samples) as training, validation, and testing sets respectively. The experimental data setup is illustrated in Fig.~\ref{fig:datasets_configuration}. This experiment is repeated 5 times to study the statistical performance.

For training, we use the Adam optimizer with a learning rate of $10^{-3}$. The batch size is set to 64. If the validation accuracy does not improve over 5 consecutive epochs, the learning rate is reduced by a factor of 10. If the validation accuracy does not improve over 10 consecutive epochs, an early stopping is triggered to terminate the training process. The model that yields the highest accuracy on the validation set is kept for testing. 

\section{Results}
\label{sec:results}
\begin{figure}[!htp]
    \centering
    \includegraphics[width=1\linewidth]{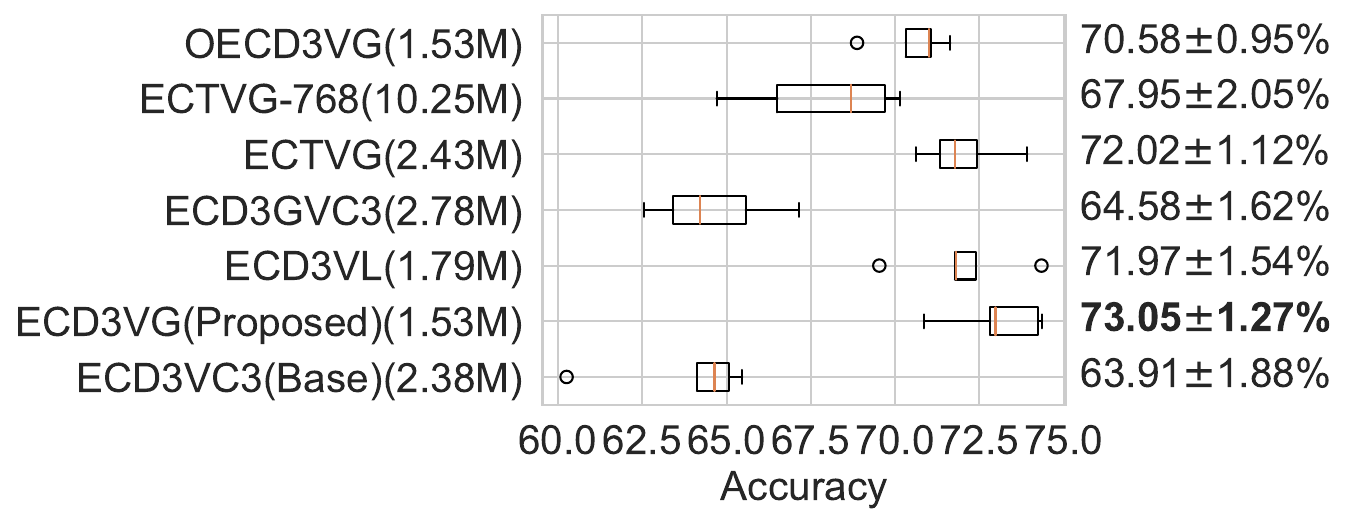}
    \caption{Performance of the proposed and compared models. The number given in each pair of parentheses denotes the number of parameters in the corresponding model.}
    \label{fig:model_performance}
\end{figure}

The results of the studied models are shown in Fig.~\ref{fig:model_performance} where E and V denote, respectively, EEG and video branches, and C, D, T, G, L, and O stand, respectively, for CNN, DCNN, TRFM, GRU, LSTM and the one-way baseline model. For example, ``ECD3VG'' denotes that the EEG branch consists of 1 convolutional layer and 3 dilated convolutional layers in sequence and the video branch has 1 GRU layer. The feature dimension for all the models is except for the model with the ``-768'' extension in which the dimension is 768. 

From Fig.~\ref{fig:model_performance}, one can clearly see that the models with GRU/LSTM module in the video branch outperform the CNN-based module, indicating that using sequence models helps capture the contextual information of video signals. Complex models with a large number of trainable parameters tend to exhibit suboptimal performance. The underlying reason is attributed to the scarcity of training samples. The best-performing model, which leverages the recurrent and convolutional layers, outperforms both baseline models, indicating that the proposed model can capture the association between video stimuli and corresponding neural responses. 

\begin{figure}[!htp]
    \centering
    \includegraphics[width=0.7\linewidth]{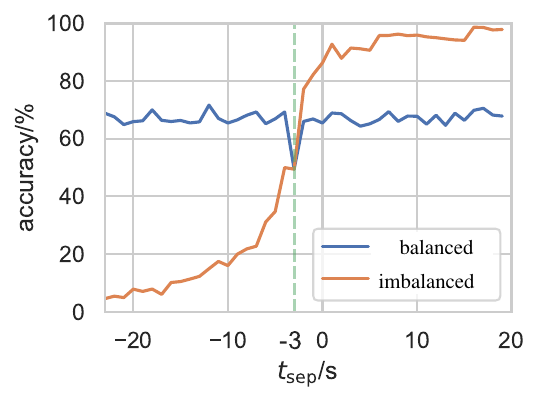}
    \caption{Accuracy of the balanced and imbalanced models on differentiating matching samples from mismatching samples extracted at different time offsets. The x-axis shows the time offset $t_{\mathrm{sep}}$ and the y-axis shows the accuracy}
    \label{fig:imbalance and balance acc_with_separation_secs}
\end{figure}

In this set of experiments, we investigate the benefits of using both the positively and negatively shifted samples as imposter samples (balanced), in comparison with using only positively shifted samples (imbalanced) as in \cite{match_mismatch_speech_eeg_dcnn_accou2021modeling}, for the EEG-video classification task. We train our model under balanced and imbalanced configurations respectively, and evaluate the accuracy of these two models in distinguishing matching videos from mismatching (imposter) ones created with different time offsets ($t_{\mathrm{sep}}$), as shown in Fig.~\ref{fig:imbalance and balance acc_with_separation_secs}. As expected, the accuracy drops to approximately 50\% at $t_{\mathrm{sep}}=-3\mathrm{s}$ where the matching and mismatching videos are identical. The accuracy of our proposed model (balanced) remains relatively stable at other offsets, indicating that the proposed model can distinguish matching and mismatching videos extracted at different intervals without requiring explicit training on those mismatching samples. We observe that when the training dataset is imbalanced, the accuracy on negative side imposter samples drops below 20\%. This suggests that the model simply memorizes the order of training examples and predicts the relatively earlier video as the matching one.

\begin{figure}[!htp]
    \centering
    \includegraphics[width=0.9\linewidth]{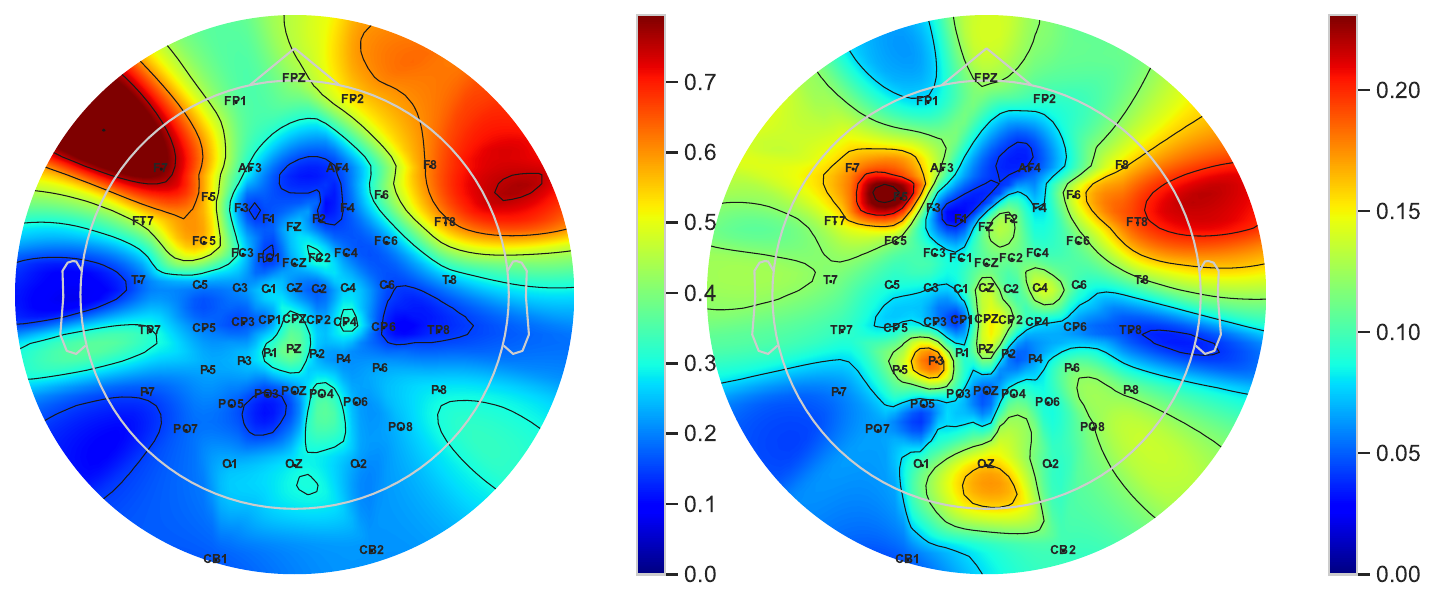}
    \caption{The topographic map: the mean (left) and the standard deviation (right) of the Grad-CAM activation scores for all EEG channels.}
    \label{fig:topo_map_head_attention}
\end{figure}

To further study which brain region contributes more to model predictions, we compute and visualize the Gradient-weighted Class Activation Mapping (Grad-CAM) scores of the proposed model \cite{grad_cam_selvaraju2016grad}. The proposed model is trained 5 times and we compute activation scores for these models. The mean and standard deviation of the scores are shown in Fig.~\ref{fig:topo_map_head_attention}. One can see from the results that the Oz area (around the visual cortex responsible for vision processing) and the Pz area (the parietal lobe related to information fusion) are highlighted in the activation map with a score of around 0.4. Furthermore, the inferior frontal gyrus area (F7 and F8), partly responsible for language processing and comprehension \cite{IFG_gao2020right}, exhibits the highest activation score (approximately 0.8), which indicates that the brain's semantic processing activities are more informative in decoding the representations of visual content in the human brain than visual processing activities \cite{brainVL_du2023decoding}.

\begin{figure}[!htp]
    \centering
    \subfloat[]
    {\includegraphics[height=0.4\linewidth]{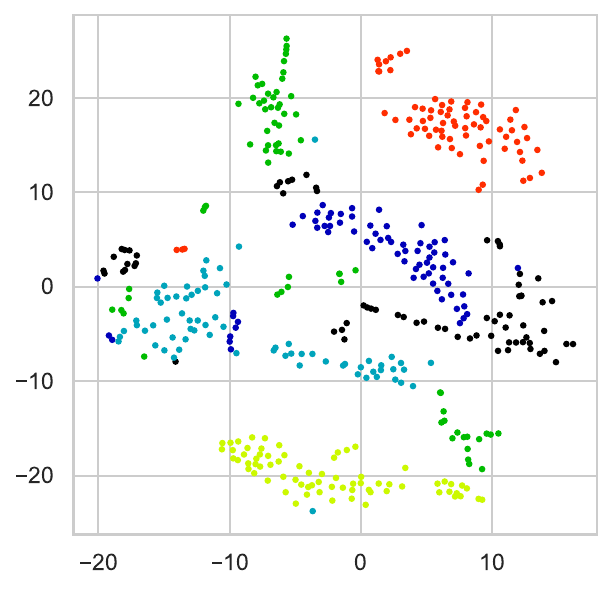}
    \label{fig:traditional_person_feature_cluster_2}}
    \subfloat[]
    {\includegraphics[height=0.4\linewidth]{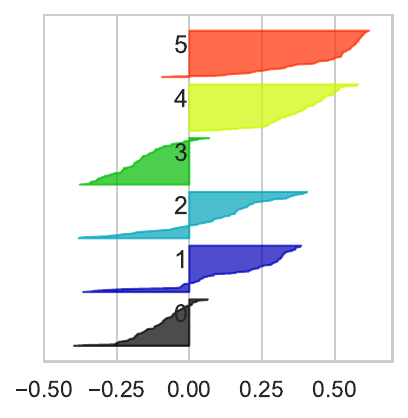}
    \label{fig:traditional_person_feature_cluster_1}}
    
    \subfloat[]
    {\includegraphics[height=0.4\linewidth]{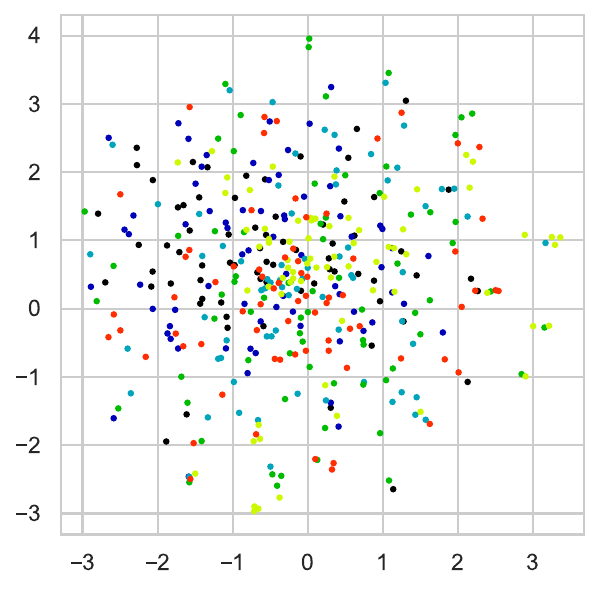}
    \label{fig:our_person_feature_cluster_2}}
    \subfloat[]
    {\includegraphics[height=0.4\linewidth]{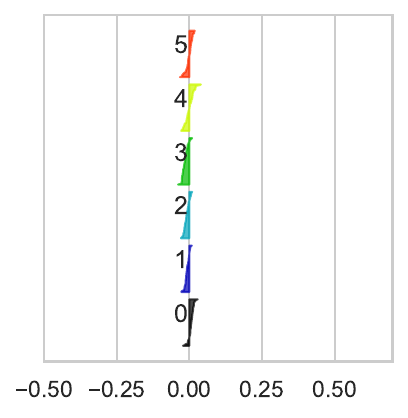}
    \label{fig:our_person_feature_cluster_1}}
    \caption{The t-SNE visualization in 2D (a, c) and silhouette analysis (b, d) of the traditional and proposed deep feature vectors. Each point represents a three-second segment of the EEG signal from one of the six testing subjects.
    }
    \label{fig:person_feature_cluster}
\end{figure}

We now examine the potential benefit of our model in terms of handling the inter-subject noise, i.e.,  how well the model is able to generalize to the EEG signals from unseen subjects. Traditional features of the EEG signals, including Hjorth parameters, differential entropy, asymmetry coefficient feature and fraction dimension mentioned in \cite{traditional_features_review}, are extracted for 6 EEG bands, namely: $\delta$ (1-3Hz), $\theta$ (4-7Hz), $\alpha$ (8-13Hz), $\beta$ (14-30Hz), $\gamma$(31-50Hz), and high-$\gamma$ (51-100Hz), which exhibits large inter-subject level variance. As shown in Fig.~\ref{fig:traditional_person_feature_cluster_2}, the EEG signals from the same subject cluster closely in the traditional feature space and are quite separable from signals from other subjects. We calculate the silhouette score over all samples using the feature vectors and the subject IDs to quantify the degree to which the undesired subject-level variance is retained. Fig.~\ref{fig:traditional_person_feature_cluster_1} shows the silhouette coefficient per subject for traditional features. Note that a higher score indicates that the feature preserves more inter-subject information. 

    
    

We then extract the latent representation of the EEG signals from the proposed model and compute silhouette scores in the embedding space of the proposed one-way baseline model for simplicity. Briefly, we extract the output of the EEG branch of the model and flatten it into a vector as the deep representation of the input EEG signal. As shown in Fig.~\ref{fig:person_feature_cluster} (b,d), the resulting embedding vectors of our model do not exhibit subject-based clusters and achieve a lower silhouette score of -0.004, while the silhouette score is 0.136 for the traditional method. This result indicates that our model is able to mitigate the inter-subject noise in the EEG signal that the traditional method is not able to handle properly while preserving the relevant neurological information encoding visual stimuli (as evidenced by the high classification accuracy). 

\FloatBarrier
\section{Conclusion}
In this work, we developed a ``match-vs-mismatch'' classification framework to model the associations between visual content and brain responses. Compared with the studied baseline methods, the proposed model achieved the highest accuracy of 73.05\% on the experimental dataset. To analyze the performance, we extracted deep embedding vectors and showed that deep representations of EEG signals have a lower subject-level silhouette score as compared to the traditional feature vectors. These experimental results suggest that the proposed deep learning model is able to effectively extract meaningful neurological features and improve the prediction of visual stimuli while suppressing the inter-subject noise. Visualizations of model activation scores reveal that brain regions associated with language and visual processing were important for model predictions. 

\vfill\pagebreak

\bibliographystyle{IEEEbib}
\bibliography{strings,refs}
\end{document}